\documentclass[letterpaper, 10 pt, conference]{ieeeconf}  

\IEEEoverridecommandlockouts                              

\usepackage[utf8]{inputenc}
\usepackage{cite}
\usepackage{amsmath,amssymb,amsfonts}
\usepackage{algorithmic}
\usepackage{graphicx}
\usepackage{textcomp}
\usepackage{xcolor}
\usepackage{breqn}
\usepackage[ruled,vlined,linesnumbered]{algorithm2e}
\usepackage{epstopdf}
\setkeys{breqn}{breakdepth={1}}

\usepackage{ntheorem}
\newtheorem*{problem}{Problem}

\title{Linear Time-Varying MPC for Nonprehensile Object Manipulation with a Nonholonomic Mobile Robot}

\author{Filippo Bertoncelli$^1$, Fabio Ruggiero$^2$, Lorenzo Sabattini$^1$
\thanks{
 The research leading to these results has been partially supported by the WELDON project, in the frame of Programme STAR, financially supported by UniNA and Compagnia di San Paolo. The authors are solely responsible for its content.
}%
\thanks{Authors would like to thank Dr. L. Fontanili, Prof. M. Milani and the Hydraulic System Design Research Group at the University of Modena and Reggio Emilia, for their support in the development of the experiments.}
\thanks{$^1$ Authors are with the Department of Sciences and Methods for Engineering (DISMI), University of Modena and Reggio Emilia, Italy.}
\thanks{$^2$ Author is with the PRISMA Lab, Department of Electrical Engineering and Information Technology, University of Naples, Via Claudio 21, 80125, Naples, Italy.}%
\thanks{Corresponding author's email {\small filippo.bertoncelli@unimore.it}}%
}

\begin{document}
\setlength{\textfloatsep}{0.5\baselineskip plus 1.5\baselineskip minus 0.9\baselineskip}
\maketitle
\begin{abstract}
This paper proposes a technique to manipulate an object with a nonholonomic mobile robot by pushing, which is a nonprehensile manipulation motion primitive. Such a primitive involves unilateral constraints associated with the friction between the robot and the manipulated object. Violating this constraint produces the slippage of the object during the manipulation, preventing the correct achievement of the task. A linear time-varying model predictive control is designed to include the unilateral constraint within the control action properly. The approach is verified in a dynamic simulation environment through a Pioneer 3-DX wheeled robot executing the pushing manipulation of a package.
\end{abstract}

\section{Introduction}

In a robotic nonprehensile manipulation task, the object is subject only to unilateral constraints imposed by both the robot manipulating it and the environment. 
A complicated manipulation task can be split into many simpler subtasks, usually called \emph{manipulation primitives}~\cite{Ruggiero18}.
Among these primitives, the \emph{pushing} operation is a simple solution, also adopted by humans, in those situations where the size of the manipulated object prevents an easy grasp by a gripper, or it is too heavy to be dexterously handled. 
Manipulation by pushing is intuitively simple, but it sets interesting control problems originated by the presence of the friction forces, which are complex to accurately model and causing an unpredictability of the object's motion~\cite{lynch1996}.

This paper tackles the problem of a nonholonomic mobile robot pushing an object in the environment.
The use of a mobile robot for pushing manipulation is justified to overcome the physical limits imposed by a static manipulator's workspace, or when the object is too large and/or too heavy to be grasped by a mobile manipulator with a gripper.
Practical applications can be primarily found in warehouses and industries for handling of goods~\cite{ram2017}.
The pursued approach is the design of a linear time-varying (LTV) model predictive control (MPC)~\cite{LTVMPC} which explicitly includes the pushing constraints. 
Violating these constraints means that the forces exerted by the robot on the object do not belong to the friction cones at the contact points. This induces the slipping of the object, reduces the precision of the manipulation task, and worsens the overall performance. 

\begin{figure}[t]
    \centering
    \includegraphics[width=0.9\linewidth]{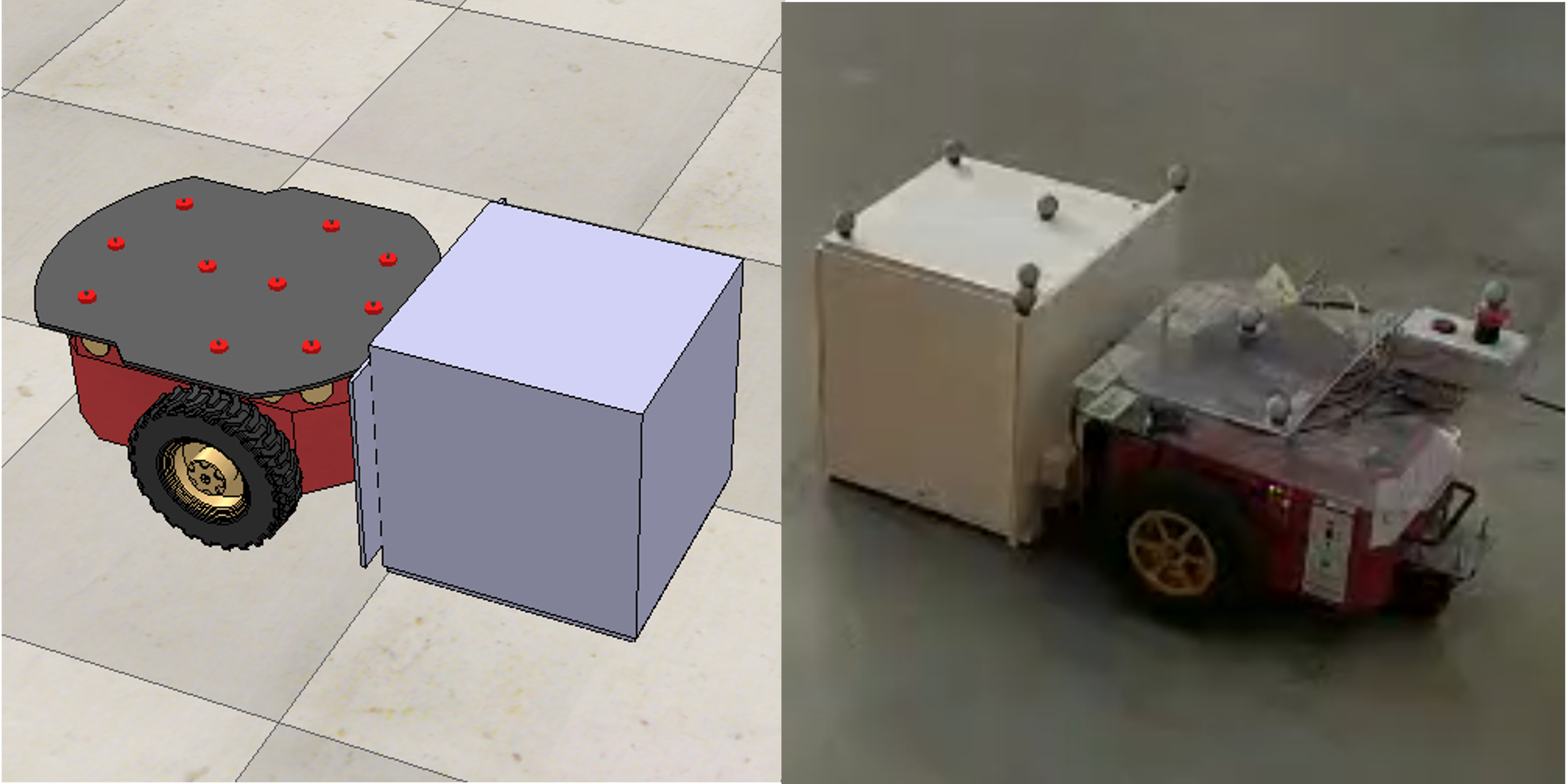}
    \caption{Example of a pushing manipulation through a mobile robot. On the left, the simulation environment. On the right, the robot in action during the carried out experiments.}
    \label{fig:setup}
\end{figure}

Building upon~\cite{Hogan2017ReactivePM}, we propose the use of a classic mobile robot to perform a pushing operation (see Fig.~\ref{fig:setup}), by considering the robot's nonholonomic constraints in the controller formulation explicitly, and introducing a motion constraint that considers the pushing dynamics to maintain a stiff contact between the robot and the pushed object.



\section{Related work}\label{sec:related}
A thorough literature review reveals that object manipulation with a mobile robot is typically achieved with proper tools~\cite{wang2016}, like grippers, or with multi-robot systems caging the object~\cite{yamashita2003}.
This latter approach takes inspiration from the natural world. Small animals, like ants, collaborate to transport heavy and large loads: several works try to mimic the behavior of ants to achieve collaborative transportation for groups of mobile robots~\cite{mccreery2014,ohashi2014}. However, this problem is often solved considering approaches in which force closure is achieved~\cite{Prattichizzo16}, planning the motion of the robots in such a way that they are opportunely displaced around the object~\cite{yamashita2003}. As an example, robots can be controlled to create a formation around the objects, in such a way that, locally exchanging information, they can transport it as designed in~\cite{fink2008}. Along the same lines, the presence of a large number of robots, that can be attached to the object in such a way that they can exchange a force with the object itself, is considered in~\cite{habibi2015}. The center of mass of the object is approximated as the centroid of the positions of the robots. Geometrical properties of the object are also estimated in~\cite{wang2002} based on the robots' relative positions.

The approaches mentioned above resemble the most common solution exploited in robotics for solving the problem of moving an object: the pick-and-place method, where the object is grasped stiffly and is then moved to the desired location. While pick-and-place is a common and effective solution in several cases, it cannot always be applied. This is particularly true when the size of the object is too large, when its shape is unknown a priori, when it is excessively heavy, when a firm grasp can damage its surface, or when the external environment places some constraints on the use of a multi-robot system. 
Nonprehensile manipulation approaches can thus be exploited in these cases. Specifically, these approaches include methods in which the robot imposes the object's motion through unilateral constraints only, such as in the case of pushing. 
The advantage is the possibility of using only one robot for the operation and the possibility of breaking a contact and create new ones during the same task~\cite{lynch1996,Ruggiero18}.
Nevertheless, nonprehensile manipulation requires taking into account the robot, the object, and the environmental dynamics, which is often a nontrivial task in the presence of the friction, as for pushing.

Recent implementations of nonprehensile manipulation with robots saw the use of flexible elements like ropes and cables.
A robot equipped with a flexible cable is shown in~\cite{kim2017}, where a planning method is proposed to exploit the cable for moving the object.
Objects with general shape are instead addressed in~\cite{maneewarn2005}, where two mobile robots are connected through a cable, and they cooperatively pull a heavy object. However, such physical interconnection between the two robots may significantly limit the freedom of motion.





To avoid these issues, mobile robots can directly perform nonprehensile manipulation by directly pushing the object~\cite{kolhe2010}.
However, it is necessary to guarantee the possibility for the mobile robot to change the pushing direction. This means that the robot must freely move in the environment, without hitting obstacles, to change its relative position to the object.
Uncertainties in control and motion execution are addressed in~\cite{krivic2016} employing an appropriate motion planning strategy, that considers an increased size of the pushed object, to include repositioning maneuvers of the pushing robot. A reinforcement learning framework is proposed in~\cite{kovac2004} to define the motion pattern for two robots pushing a box. However, a very simplistic scenario is considered where dynamics are neglected. By measuring the instantaneous direction, a robot or a group of robots is guided by an artificial potential field in~\cite{igarashi2010} to push an object. Also in this case, dynamic effects, such as friction, are not considered, making the proposed method unsuitable for complex situations, such as in the presence of non-uniform friction. A fuzzy controller is instead designed in~\cite{golkar2009} to control two robots pushing an object with known geometrical properties.
Slipping of the mobile robot's wheels during a pushing operation is avoided in~\cite{Bertoncelli19} through a nonlinear MPC design. However, the friction between the manipulated object and the robot is neglected, causing possible slippage of the object during the manipulation task.
Finally, the complexity of determining the optimal sequence of actions to manipulate an object by pushing is solved offline in~\cite{Hogan2017ReactivePM} through machine learning. A convex hybrid MPC program is then solved online to achieve planar manipulation.

Differently from~\cite{Hogan2017ReactivePM}, the proposed design explicitly includes the nonholonomic nature of the most common mobile robots available in the market with the dynamic model formulation. Besides, a motion constraint is designed to avoid the slippage of the contact with pushed object during the manipulation, made necessary by the robot nonholonomy.


\section{Problem statement}\label{sec:prob}

Consider a wheeled mobile robot moving in a bi-dimensional environment, where a polygonal planar object has to be manipulated.
Let $\Sigma_w$ and $\Sigma_r$ be the global reference frame and the body frame attached to the center of the robot's axle, respectively.
Let the pose of the mobile robot at time $t$ be represented by the vector $\chi_r(t) = \begin{bmatrix}x_r(t)& y_r(t)& \theta_r(t)\end{bmatrix}^T\in\mathbb{R}^3$, where $x_r(t),y_r(t) \in\mathbb{R}$ represent the position of $\Sigma_r$ in $\Sigma_w$, and $\theta_r \in \mathbb{R}$ is the rotation of $\Sigma_r$ with respect to $\Sigma_w$. 
A visualization is provided in Fig.~\ref{fig:ddmr}. In a similar way, let $\chi_o(t) = \begin{bmatrix}x_o(t)& y_o(t)& \theta_o(t)\end{bmatrix}^T\in\mathbb{R}^3$ represent the pose of the object, as shown in Fig.~\ref{fig:contactforces}.

We provide a solution to the following problem.
\begin{problem}
Control the motion of the mobile robot to manipulate the object through pushing maneuvers, in such a way that the trajectory $\chi_o(t)$ tracks the desired one $\chi_d(t)$ with the desired accuracy, starting from the initial pose $\chi_o(0)$.
\end{problem}
In the following, we will assume the mobile robot to behave as a unicycle. The choice is motivated by the simplicity of notation introduced by such a model, and by the fact that several real-world mobile robots (as differential-drive robots) can be represented according to this formulation~\cite{oriolo2002}. 
We assume that all the considered contacts are rigid, that the robot wheels do not slip, and that the forces exchanged in the interaction follow Coulomb's model of friction. Moreover, we decompose each contact force in two components, aligned with the edges of the friction cone\cite{Prattichizzo2008}. 
The angle between each component and the contact normal is
\begin{equation}
\theta_\mu= \tan^{-1}\mu,
\label{eq:thetamu}
\end{equation}
where 
%
$\mu > 0$ is the friction coefficient associated with the interacting surfaces. A visualization of the used decomposition is provided in Figure~\ref{fig:contactforces}.
The motion of the controlled system is assumed quasistatic (i.e, it is slow enough that inertial forces are negligible).
Moreover, we assume the mobile robot to be equipped with a planar end-effector (i.e., a planar contact surface), such that the surface used to interact with the object is consistent and homogeneous. During the interaction, the end-effector is supposed to be parallel to one of the sides of the polygonal object. This type of interaction is typically referred to as 
\emph{line contact}, modeled as if the only contact points were the extreme points of the line~\cite{xie_linecontact}.

\section{Modeling}
In this section, we introduce the mathematical model of the system motion and the constraints applied within the MPC controller for planar manipulation tasks. First, we describe a second-order model of the mobile robot and its error dynamic for the desired trajectory, then the mathematical model of the pushed object motion is introduced.
\subsection{Robot Model}
\begin{figure}[t]
    \centering
    \includegraphics[width=0.9\linewidth]{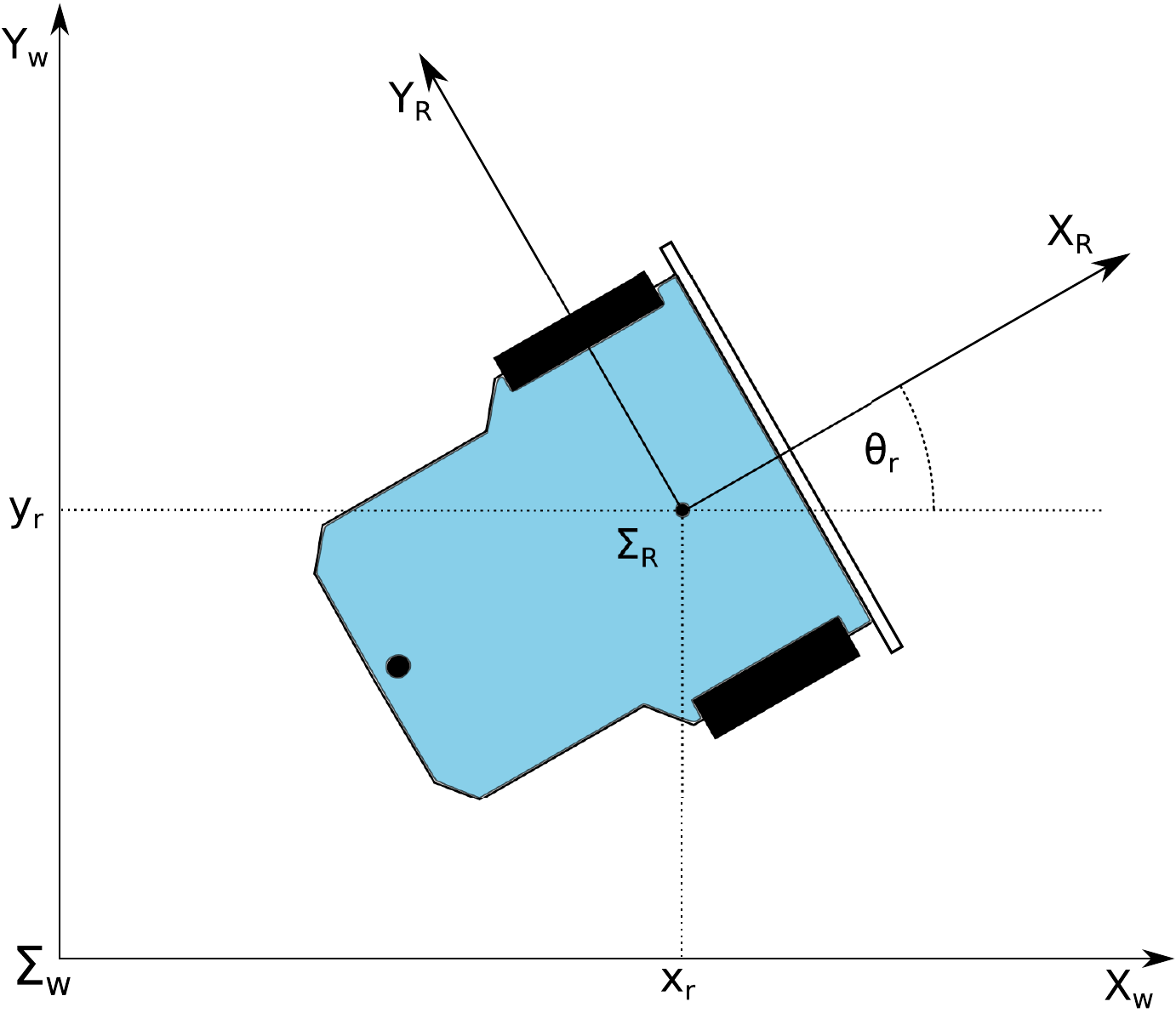}
    \caption{Schematic representation of the differential drive mobile robot. The black rectangles are the wheels. The black circle is the caster wheel. The frontal bumper is represented by the white rectangle in front of the wheels.}
    \label{fig:ddmr}
\end{figure}

Defining $v_r,\omega_r\in\mathbb{R}$ as the linear and angular velocities of the robot, respectively (see Fig.~\ref{fig:ddmr}), the state of the robot is defined as $\xi(t)\in\mathbb{R}^5$, given by
\begin{equation}
\xi(t)=\begin{bmatrix}x_r(t)&y_r(t)&\theta_r(t)&v_r(t)&\omega_r(t)\end{bmatrix}^T.
\label{eq:state}
\end{equation}
For ease of notation, in the following, dependence on time will be omitted, when not strictly necessary.

Define now $a_r, \varepsilon_r\in\mathbb{R}$ as the inputs for the robot, given as the linear and angular acceleration, respectively. Hence, the model of the robot motion can be written as
\begin{align}
    \dot{\xi}&=\begin{bmatrix}\cos\theta_r v_r\\\sin\theta_r v_r\\\omega_r\\0\\0\end{bmatrix}+\begin{bmatrix}0\\0\\0\\a_r\\\varepsilon_r\end{bmatrix}.
    \label{eq:motionmodel}
\end{align}
Solution of the Problem stated in Section~\ref{sec:prob} passes through the generation of the desired trajectory $\xi_d(t)=\begin{bmatrix}x_{rd}(t)& y_{rd}(t)& \theta_{rd}(t)& v_{rd}(t)& \omega_{rd}(t)\end{bmatrix}^T\in\mathbb{R}^5$ for the robot to realize the pushing maneuvers\footnote{Several strategies exist, in the literature, to generate trajectories for mobile robots during pushing maneuvers. Due to space limitations, this problem is not addressed in this paper. However, the reader is referred to, e.g., \cite{lynch1996}\cite{WoodruffLynch2017}, for further details.}.

Let ${e}(t)=
    \begin{bmatrix}
        e_{xr}(t)&e_{yr}(t)&e_{\theta r}(t)&e_{vr}(t)&e_{\omega r}(t)
\end{bmatrix}^T \in \mathbb{R}^5$ be the error vector with respect to the desired reference frame centered in $\left(x_{rd}(t),\, y_{rd}(t)\right)$, and oriented as $\theta_{rd}(t)$, that is defined as
%

\begin{align}
    {e}(t)=\begin{bmatrix}\cos(\theta_{rd}(t))&\sin(\theta_{rd}(t))&0&0&0\\
    -\sin(\theta_{rd}(t))&\cos(\theta_{rd}(t))&0&0&0\\
    0&0&1&0&0\\
    0&0&0&1&0\\
    0&0&0&0&1\end{bmatrix}(\xi -\xi_d).
    \label{eq:errordef}
\end{align}
Considering the robot motion~\eqref{eq:motionmodel} and the error vector~\eqref{eq:errordef}, we can describe the system error dynamics as follows:
\begin{align}
    \label{eq:emodel}
    {\dot{e}}(t)=f(t)=\begin{bmatrix}\cos(e_{\theta r})(e_{vr} + v_{rd}) - v_{rd} + e_{yr}\omega_{rd}\\
    \sin(e_{\theta r})(e_{vr} + v_{rd}) - e_{xr}\omega_{rd}\\
     e_{\omega r}\\
     a_r - \dot{v}_{rd}\\
     \varepsilon_r - \dot{\omega}_{rd}
    \end{bmatrix}.
\end{align}

\subsection{Pushed Object Model}\label{sec:pushobjmot}
\begin{figure}[t]
    \centering
    \includegraphics[width=0.9\linewidth]{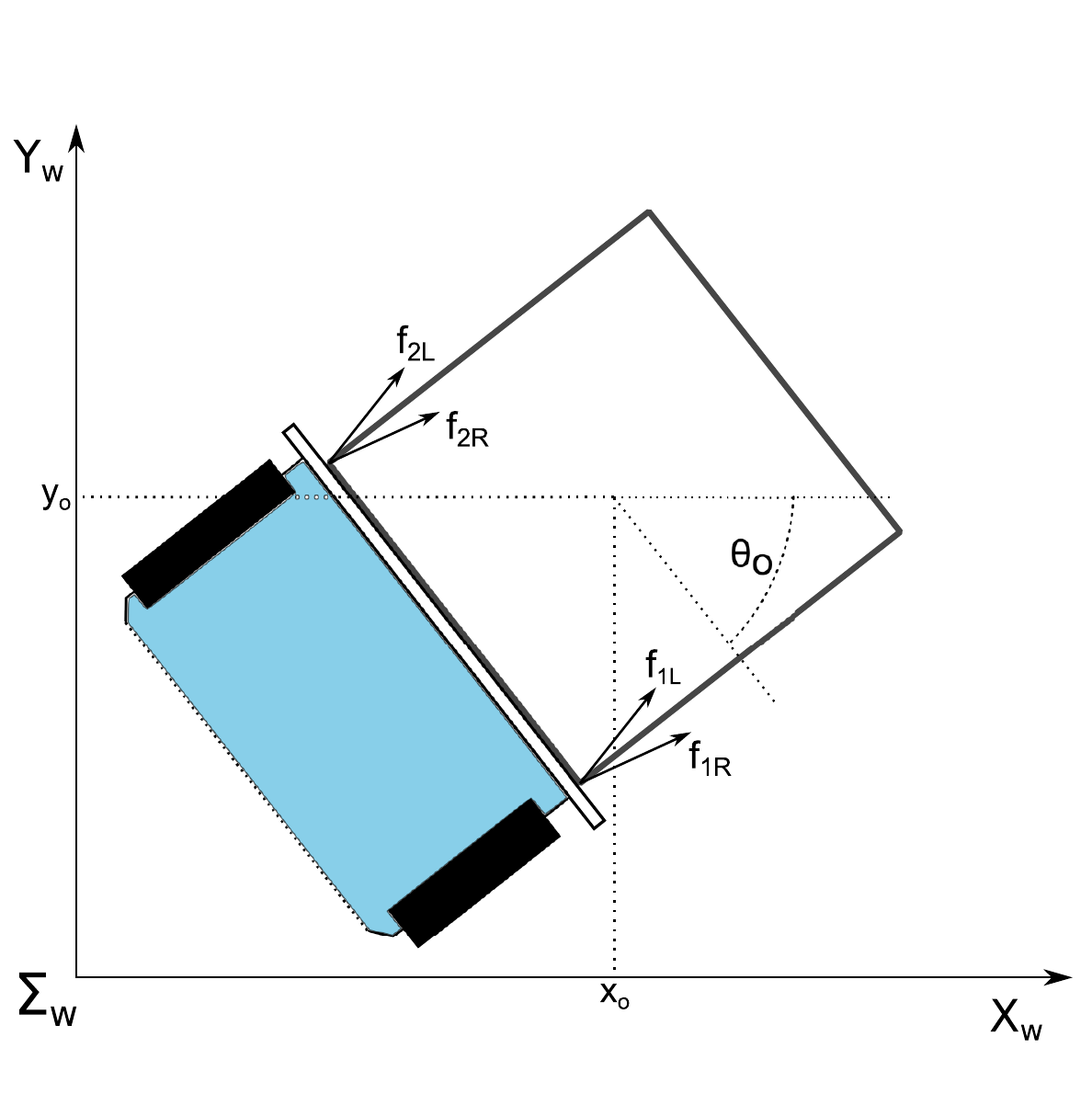}
    \caption{Schematic representation of the pushed object.}
    \label{fig:contactforces}
\end{figure}

Consider, as discussed in Section~\ref{sec:prob}, a polygonal object pushed by the robot with line contact on one of its sides. The contact forces are modeled using the components along the friction cone as shown in Fig.~\ref{fig:contactforces}. We denote with $f_{iR} \in \mathbb{R}$ and $f_{iL} \in \mathbb{R}$ the right and left contact force components, respectively, for each contact point $i=\{1,2\}$\cite{Prattichizzo2008}. We define the vector $f_c\in \mathbb{R}^4$ as
\begin{align}
    \label{eq:fc}
    f_c = \begin{bmatrix}f_{1R} & f_{1L} & f_{2R} & f_{2L} \end{bmatrix}^T.
\end{align}
The total external wrench $w \in \mathbb{R}^3$, expressed in $\Sigma_W$ and whose torque is applied around the object's center of mass, exerted by the robot to the object can be described by
\begin{align}
    {w}=Gf_c
\end{align}
where $G \in \mathbb{R}^{3 \times 4}$ is the so called grasp matrix. For a square object of side length $2s>0$, the grasp matrix is
\begin{align}
    G &=\begin{bmatrix}
    \cos(\theta_r - \theta_\mu) & \sin(\theta_r - \theta_\mu) & s(\cos{\theta_\mu} + \sin{\theta_\mu})\\
    \cos(\theta_r + \theta_\mu) & \sin(\theta_r + \theta_\mu) &  s(\cos{\theta_\mu} - \sin{\theta_\mu})\\
    \cos(\theta_r - \theta_\mu) & \sin(\theta_r - \theta_\mu)& s(\sin{\theta_\mu} - \cos{\theta_\mu})\\
    \cos(\theta_r + \theta_\mu) & \sin(\theta_r + \theta_\mu) & -s(\cos{\theta_\mu} + \sin{\theta_\mu})\end{bmatrix}^T,
\end{align}
where $\theta_\mu$ is given in~\eqref{eq:thetamu}, and it can be obtained through a geometrical analysis of the contact forces.

Under the assumption of quasistatic interaction, the object motion can be described using the limit surface~\cite{Goyal1991PlanarSW}, a geometric representation of the relationship between the applied force on an object and its instantaneous velocity. Inspired by~\cite{Hogan2017ReactivePM}, an ellipsoidal approximation of the limit surface is used, due to its simplicity and invertibility properties. A convex quadratic formulation of the ellipsoidal limit surface is given by ${S}({w})=\frac{1}{2}{w}^TH{w}$, where $H \in \mathbb{R}^{3 \times 3}$ is the matrix representing the ellipsoidal approximation of the limit surface. A procedure for computing such an approximation, that requires the knowledge of the object's shape and mass as well as the friction coefficient of the support surface, can be found in \cite{Lee1991FixturePW}. Through the principle of maximal dissipation \cite{Goyal1991PlanarSW}, the object instantaneous velocity is perpendicular to the limit surface for a given wrench, which implies:
\begin{align}
    \begin{bmatrix}\dot{x}_o & \dot{y}_o & \dot{\theta}_o\end{bmatrix}^T=H{w}.
\end{align}

\section{MPC Controller Formulation}
To correctly solve the Problem defined in Section~\ref{sec:prob}, a controller must be designed such as the error vector $e(t)$ in~\eqref{eq:errordef} is steered to zero without violating the friction constraints given by the contact between the robot and the object. This avoids the slippage of the object during the pushing manipulation. As a matter of fact, zeroing the error vector only does not imply that the object follows the desired trajectory $\chi_d(t)$.
The controller makes use of a LTV MPC formulation~\cite{LTVMPC} to solve the nonlinear control problem in real-time through the solution of a motion constrained optimization problem. First, a LTV approximation of the model is presented. The model considers the presence of the velocity and acceleration of the desired trajectory in the form of the measured disturbances $v(t)\in \mathbb{R}^4$, a vector of known but unmodifyable model inputs. 
The MPC formulation and the applied constraints are finally addressed.

\subsection{LTV Model Approximation}
As discussed in \cite{borrelli_bemporad_morari_2017}, the MPC formulation requires a discrete-time linear (or linearized) model to construct the optimization problem. Therefore, the model~\eqref{eq:emodel} is linearized and discretized. The linearization is performed around a series of predicted states $\Tilde{{e}}(t)$ obtained through numerical integration of~\eqref{eq:emodel}.  
In particular, the nonlinear error dynamics \eqref{eq:emodel} can be approximated by the following LTV system
\begin{align}
    \label{eq:emodel_lcont}
    \dot{{e}}(t) = A(\Tilde{{e}}(t),{v}(t)){e}(t) + B_u {u}(t) + B_v(\Tilde{{e}}(t)){v}(t),
\end{align}
where $u(t) \in \mathbb{R}^2$ is the model input vector, ${v}(t) \in \mathbb{R}^4$ is the vector of measured disturbances and $\Tilde{{e}}(t) \in \mathbb{R}^5$ is the predicted state. 
More specifically, we define
\begin{align}
    u(t) = \begin{bmatrix}a_r\\\varepsilon_r\end{bmatrix} &\hspace{0pt}&
    {v}(t) = \begin{bmatrix}v_{rd}\\\omega_{rd}\\\dot{v}_{rd}\\\dot{\omega}_{rd}\end{bmatrix} & \hspace{0pt} & \Tilde{{e}}(t)=\begin{bmatrix}\Tilde{e}_{xr}(t)\\\Tilde{e}_{yr}(t)\\\Tilde{e}_{\theta r}(t)\\\Tilde{e}_{vr}(t)\\\Tilde{e}_{\omega r}(t) \end{bmatrix},
\end{align}
and
\begin{equation}
\begin{array}{ll}
    A(\Tilde{{e}}(t),{v}(t))=\frac{\partial {f}}{\partial {e}}\Bigr|_{\substack{{e}(t)=\Tilde{{e}}(t)\\v(t)}}& 
    \!\!\!\!B_v(\Tilde{{e}}(t))=\frac{\partial {f}}{\partial {v}}\Bigr|_{\substack{{e}(t)=\Tilde{{e}}(t)}}\\\\
    B_u = \begin{bmatrix}0&0&0&1&0\\0&0&0&0&1\end{bmatrix}^T,
    \end{array}
\end{equation}
which represents the fact that matrices are obtained performing the linearization around $\left(\Tilde{e}(t),v(t)\right)$, both evaluated at time $t$.
%
The discrete-time equivalent model of~\eqref{eq:emodel_lcont}, defined with sampling time $T_s>0$, can then be obtained following the procedure given in~\cite{Franklin1997}.
Denoting with $k \in \mathbb{Z}$ the discrete time variable, we get
\begin{align}
    {e}[k+1] = A[k]\,{e}[k] + B_{u}[k]\,{u}[k] + B_{v}[k]\,{v}[k] ,
\end{align}
with 
\begin{align}
    A[k] &= e^{A(\Tilde{{e}}[k],{v}[k])T_s},\\ B_{u}[k]&=\int_0^{T_s}e^{A(\Tilde{{e}}[k],{v}[k])\tau}d\tau B_{u}(\Tilde{{e}}[k]),\\
    B_{v}[k]&=\int_0^{T_s}e^{A(\Tilde{{e}}[k],{v}[k])\tau}d\tau B_{v}(\Tilde{{e}}[k]).
\end{align}

\subsection{LTV MPC Formulation}\label{sec:qpprob}
The idea behind the MPC formulation is to optimize the future behaviour across a finite prediction horizon of $p$ steps.
At every discrete-time instant $k$, for a given state estimate $e[k]$, the optimal control input is computed solving the following constrained quadratic programming (QP) 
\begin{subequations}
    \begin{align}
        &\underset{z_k}{\text{min}}
        &&J(z_k,e[k])\label{eq:costfunction}\\
        &\text{s.t.}
        && M z_k < b,\label{eq:constraint1}\\
        &&& u_{m}[k+i] \leq u[k+i] \leq u_{M}[k+i], i=0\dots p-1\label{eq:constraint2}\\
        &&& \Delta u_{m}[k+i] \leq \Delta u[k+i] \leq \Delta u_{M}[k+i],\\&&& i=0\dots p-1\label{eq:constraint3}
    \end{align}
    \label{eq:optimprob}
\end{subequations}
where
\begin{equation}
    \begin{array}{r}
z_k=\left[u[k]^T\, f_c^T[k]\, u[k+1]^T\, f_c^T[k+1]\, \dots\right.  \\\left.  \dots\, u[k+p-1]^T\, f_c^T[k+p-1]\right]^T
    \end{array}
\end{equation}
is the QP decision variable containing the input vector $u[k+i]$ (that collects the linear and angular acceleration of the robots) and $f_c[k+i]$ (that collects the forces imposed on the pushed object) for $i=0,\dots, p-1$. As will be detailed in Section~\ref{sec:motionconstr}, such a definition of the decision variable allows us to consider the physical limitations of the robot inside the QP problem, even though the contact forces are not considered in the robot model.
Besides, the cost function in~\eqref{eq:optimprob} is defined as the following quadratic function
%
\begin{dmath}
    J(z_k,e[k]) = \sum_{i=0}^{p-1}\{[{e}[k+i]^T Q {e}[k+i]]+[{u}[k+i]^TR_u {u}[k+i]]+[\Delta {u}[k+i]^T R_{\Delta u} \Delta {u}[k+i]]\}+{e}[k+p]^T P {e}[k+p].
\end{dmath}
The diagonal matrices $Q,P\in \mathbb{R}^{5\times5}$ provide the weights associated with each state variable, while $R_u,R_{\Delta u}\in \mathbb{R}^{2\times2}$ contain the weights on the amplitude of the input and the amplitude of its rate of change respectively. The terminal weight $P$ is introduced to improve stability, as discussed in~\cite{Grne2013:NLMPC}. These matrices are all positive semidefinite. Inequality \eqref{eq:constraint1} expresses a pushing interaction constraint, which will be described in details in Section~\ref{sec:motionconstr}. Expression \eqref{eq:constraint2} imposes upper and lower limits on the elements of the QP decision variable $z_k$, while \eqref{eq:constraint3} sets limits on its rate of change. These constraints are imposed to guarantee the feasibility of the solution, taking into account the physical limitations of the robot actuators.

\subsection{Pushing Constraints for Object Slippage Avoidance}\label{sec:motionconstr}

Since the robot is subject to nonholonomic constraints, it cannot change its orientation instantaneously. 
The direction of the force applied to the pushed object is thus constrained as well. Besides, as previously discussed, the pushing force must be restricted within the friction cone to avoid object slippage during manipulation. 
Hence, we will now introduce a constraint for the robot motion, such that the contact between the robot and the object does not break. This allows us to guarantee that the movement of the robot produces valid pushing forces, that lie within the friction cone. As a consequence, the input for the robot does not generate any relative motion between the object and the robot itself.

The concept above is implemented imposing the following constraints
\begin{align}
\label{eq:pushingconstraint}
\begin{bmatrix}\dot{x}_r\\\dot{y}_r\\\dot{\theta}_r\end{bmatrix}+\omega_r \times R(\theta_r){p}_{or}=\begin{bmatrix}\dot{x}_o\\\dot{y}_o\\\dot{\theta}_o\end{bmatrix}=HGf_c,
\end{align} 
where ${p}_{or}\in\mathbb{R}^2$ is the position of the object in the robot frame $\Sigma_r$, $R(\theta_r)\in SO(2)$ is the rotation matrix between $\Sigma_r$ and $\Sigma_w$. 
The left-hand side of~\eqref{eq:pushingconstraint} represents the velocity that the object would have if the robot-object system were moving as a rigid body (i.e., no relative motion). The right-hand side expresses the motion of the object due to contact forces, as explained in Section~\ref{sec:pushobjmot}. 

In order to include~\eqref{eq:pushingconstraint} inside the optimization problem~\eqref{eq:optimprob}, some adjustments are required.
In particular, to ensure that the contact forces lie inside the friction cone, each component of $f_c$ is bounded to be greater than zero. 
The equality constraint in~\eqref{eq:pushingconstraint} is thus converted into a set of two double inequalities, of the form $g(z_k)\leq 0$, and linearized, at each time step $k$, around the point $(\Tilde{{e}}[k],{v}[k],{p}_{or}[k])$. 
Matrix $M\in \mathbb{R}^{6p\times 6p}$ in~\eqref{eq:constraint1} is finally defined as the Jacobian matrix of the left-hand side of the inequalities, computed with respect to variable $z_k$, while vector $b\in \mathbb{R}^{6p}$ is a zero vector.

\section{Implementation and Experiments}
In this section, we discuss the implementation of the pushing system and the results of three different manipulation tests, which are representative of different operative conditions. The robot used during the simulation is a Pioneer 3-DX, a differential drive mobile robot with two actuated wheels and a castor wheel. The robot is equipped with a pushing bumper attached on the front. The robot receives velocity commands in the form $v_r,\omega_r$ through ROS~\cite{ROS}. At each time step $k$, with period $T_s=0.1s$, the following procedure is performed. The controller first sends the velocity command to the robot, then collects the data required to predict the future behaviour and generate the linearized models. The solution of the quadratic problem discussed in section \ref{sec:qpprob} is then computed and used to generate the velocity command for the future step.  
\begin{algorithm}
\DontPrintSemicolon
send previous $(v_r,\omega_r)$\;
get ${e}[k],{v}[k]$\;
$\Tilde{{e}}[k]\xleftarrow{}$predict(${e}[k]$,${v}[k]$,$z_{k-1}$)\;
$A_k,B_{u_k},B_{v_k},M\xleftarrow{}$linearize$(\Tilde{{e}}[k],{v}[k])$\;
$z_k\xleftarrow{}$ QP($A_k,B_{u_k},B_{v_k},M,\Tilde{{e}}[k],{v}[k]$)\;
$(v_r,\omega_r)\xleftarrow{}(v_r,\omega_r)+T_s {u}[k]$\;
$z_{k-1} \xleftarrow{} z_k$
\caption{Feedback procedure}
\end{algorithm}

Two case studies have been carried out in simulations, performed on a laptop with an Intel Core i7-4510U using the CoppeliaSym physics simulator. The validated controller is written in MATLAB, while ROS handles the communication with the simulator. The gains are experimentally tuned to $Q = diag([15,20,5,1,1])$, $P=50Q$, $R_u = diag([0.1,0.1,0.001,0.001,0.001,0.001])$, and $R_{\Delta u}=diag([0.1,0.1,0,0,0,0])$. The results of the simulations are discussed below and are reported in the accompanying video. The video also reports the results of preliminary experiments, performed with a real robotic system in a laboratory environment.

\subsection{Tracking along a straight line}
\begin{figure}[t]
    \centering
    \includegraphics[width=\linewidth]{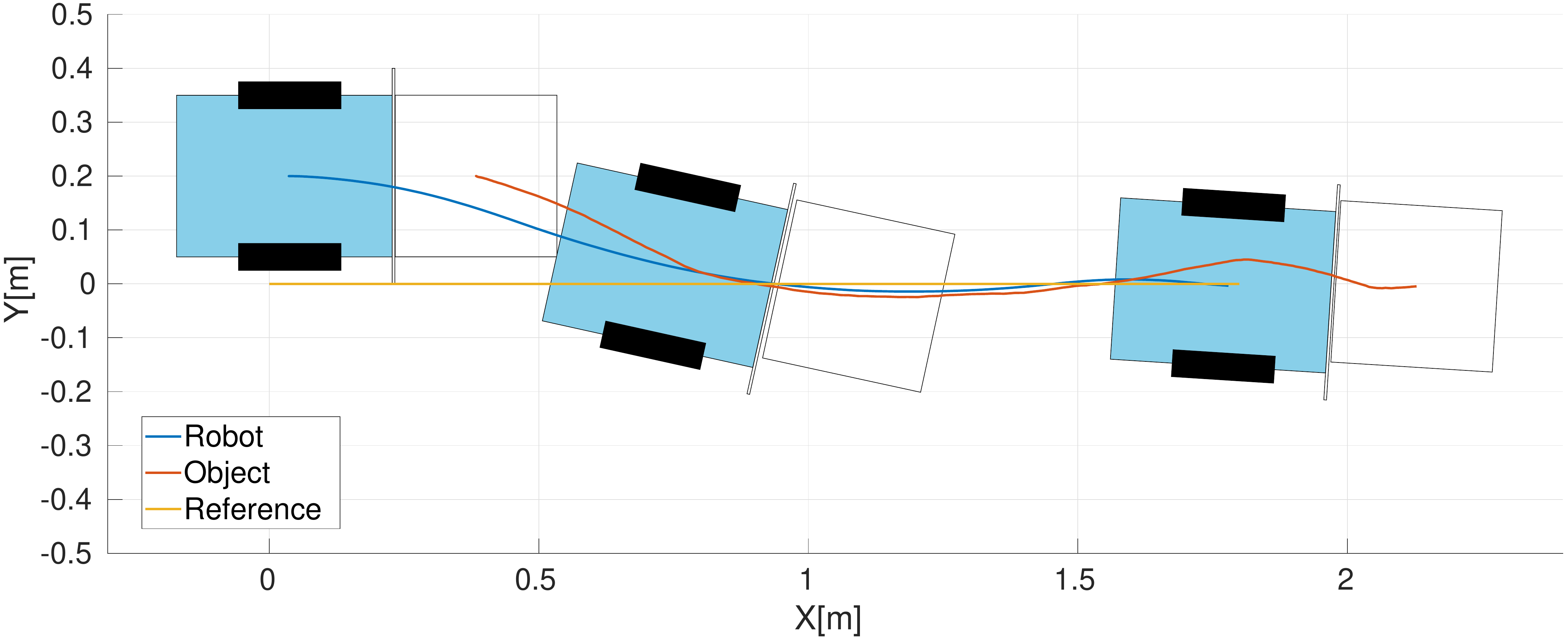}
    \caption{Line tracking from a non-zero initial error state using the proposed controller and constraints.}
    \label{fig:line2d}
\end{figure}

\begin{figure}[t]
    \centering
    \includegraphics[width=\linewidth]{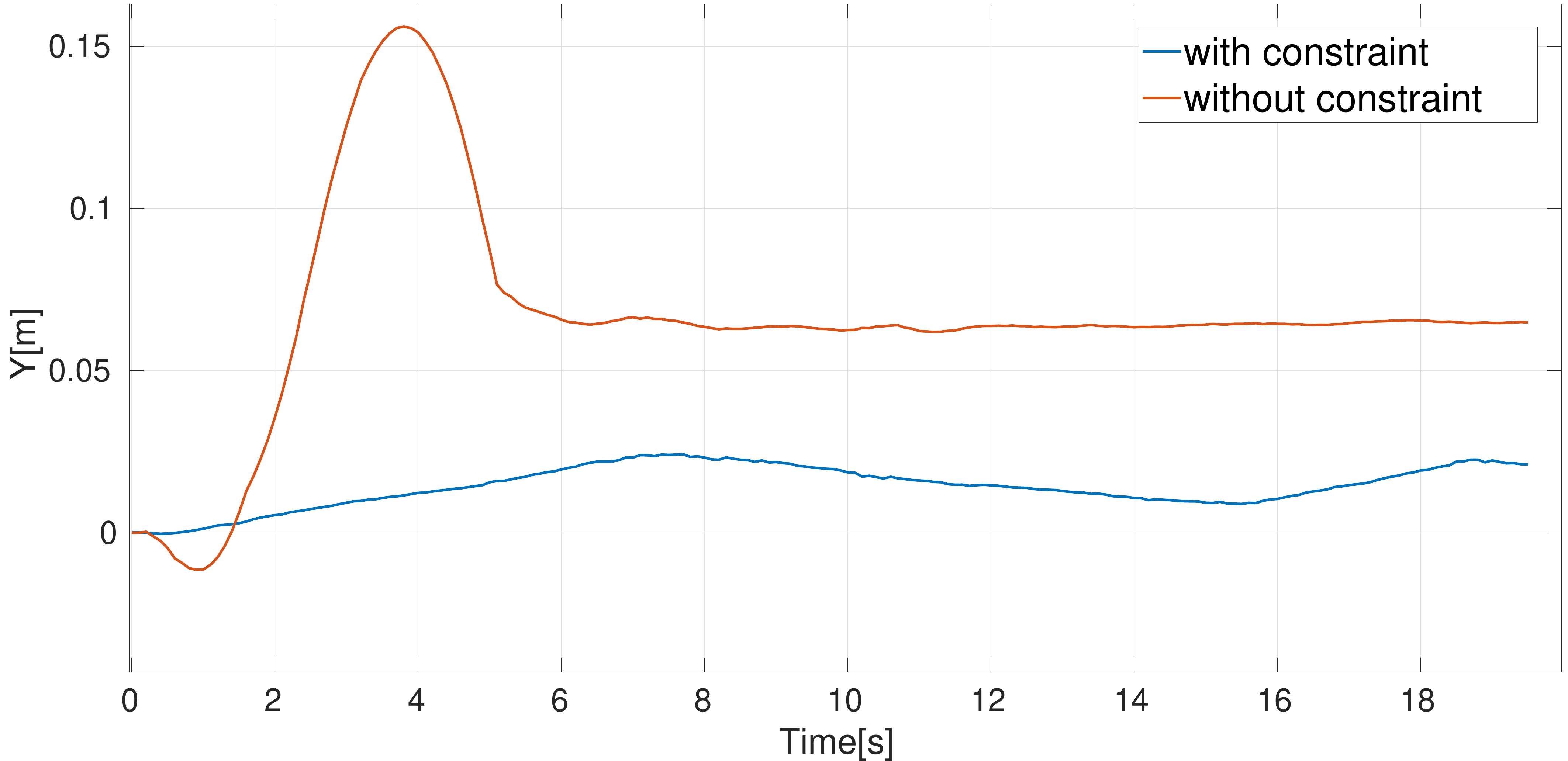}
    \caption{Y position of the pushed object with respect to $\Sigma_r$ during the manipulation.}
    \label{fig:ypos}
\end{figure}

\begin{figure}[t]
    \centering
    \includegraphics[width=\linewidth]{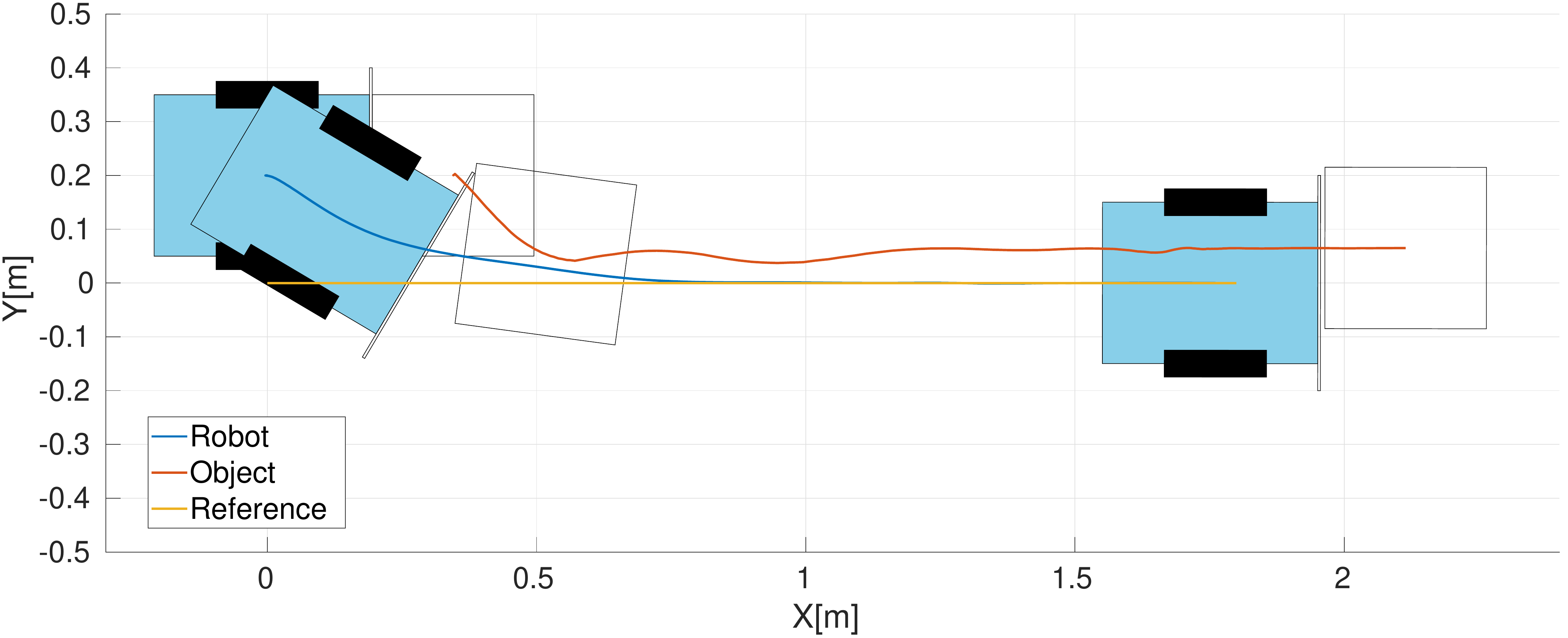}
    \caption{Line tracking from a non-zero initial error state using the controller without the proposed constraint.}
    \label{fig:line2dnoc}
\end{figure}

The first case study we propose is the tracking of a straight line starting with an offset. Several simulations have been performed, with the robot starting its movement with initial error state ${e}(0)=\begin{bmatrix}0&\varphi&0&0&0\end{bmatrix}^T$, with varying $\varphi \in \{0.1, 0.2, 0.4, 0.5\}$. A representative run of the simulations, performed with $\varphi=0.2$, is discussed hereafter.
The pushed object is placed in contact with the robot in a centered position. Figure \ref{fig:line2d} depicts the planar movement of the robot and the object, measured for a representative run of the simulations. The yellow line represents the desired trajectory while the blue line and red line depicts the movement of the robot and the object, respectively. The figure clearly shows that an initial position offset can be corrected using the proposed controller and constraints.
Figure~\ref{fig:ypos} shows a comparison of the y components of the object position with respect to $\Sigma_r$, while being pushed, with and without the presence of the constraint in the controller. The application of the constraint significantly reduces the amplitude of the movement of the pushed object.  The same conclusion can be extracted from Figure~\ref{fig:line2dnoc}, that shows the positions of the object and robot controlled without the constraint~\eqref{eq:pushingconstraint}. Moreover, Figure~\ref{fig:line2dnoc} shows that, during the manipulation, the movement of the robot causes an interruption of the pushing line contact, also reflected in the spike visualized in Figure~\ref{fig:ypos}, that results in a loss of control over the movement of the object and ultimately in a loss of quality of the manipulation. The proposed controller and constraints maintain the line contact, with a final average object position error less than $0.01$~m, while the absence of the constraint on the motion leads to an average error greater than $0.05$~m.

\subsection{Complete manipulation task}
\begin{figure}[t]
    \centering
    \includegraphics[width=\linewidth]{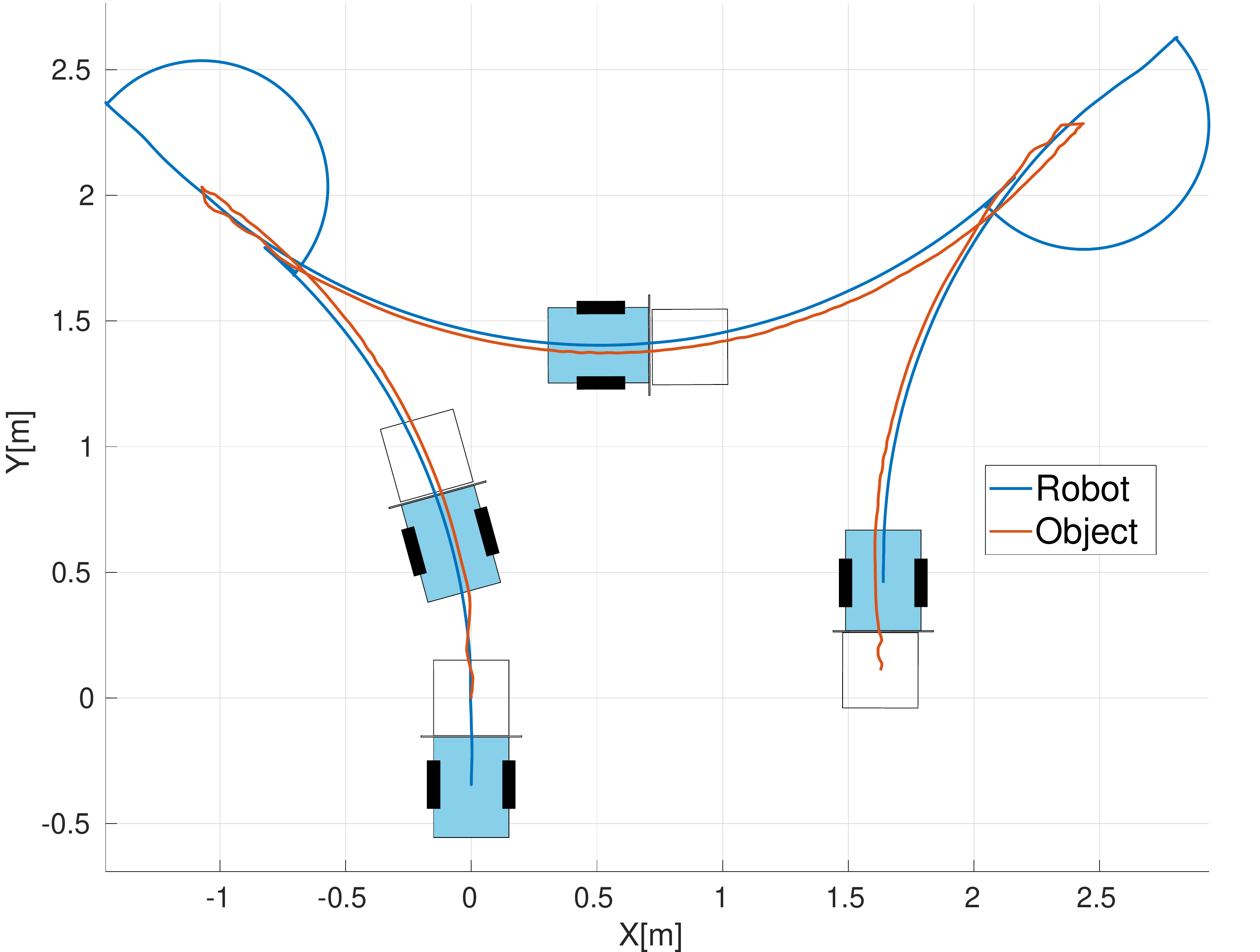}
    \caption{Complete manipulation of the object.}
    \label{fig:manip}
\end{figure}

The second case study we explore is a complete manipulation. To complete the task, the robot transports the object to a desired configuration performing a series of pushing actions. Once a pushing maneuver is completed, the robot performs a repositioning maneuver to change the pushing side before starting the next action. During the maneuver the robot steps back from the object, goes around the object along a circular trajectory and then approaches slowly the object until the contact is established. Several simulations have been performed, considering different trajectories composed of straight and curve segments. The trajectory traveled by the robot and the object during a representative run of the simulations is depicted in Figure~\ref{fig:manip}. In this task the robot transports the object from the initial position $\chi_o(0)=\begin{bmatrix}0.0& 0.0& 0.0\end{bmatrix}^T$ towards the desired configuration $\begin{bmatrix}1.65& 0.15&\pi\end{bmatrix}^T$ performing three pushing actions on the object.  The results indicate that, with the use of the proposed controller, it is possible to track a curved line to transport the package without significant accumulation of error, that implies that the robot, when an appropriate reference trajectory is provided, can manipulate the object across the environment. The final positon error is $0.04$~m while the final orientation error is $1.8$~deg.

\section{Conclusion and Future Work}
In this paper, we investigated the problem of manipulating an object in a bi-dimensional environment by pushing with a nonholonomic mobile robot. In particular, we designed a predictive controller for the mobile robot with an appropriate set of constraints to ensure the correct manipulation, providing stiff contact with the object.
Numerical case studies are presented, while early-stage experiments are shown in the multimedia attachment.

Future work is focused on consolidating the experimental validation of the proposed approach, taking into account possible external disturbances. 
We would also like to include our previous work developed in~\cite{Bertoncelli19} within the proposed framework. 
Besides, we would like to extend this work to the case of a multi-robot system. Differently from what presented in Section~\ref{sec:related}, the multi-robot system should not resemble a pick-and-place operation, but each agent must perform nonprehensile manipulation through pushing.  

\bibliographystyle{IEEEtran} 
\bibliography{biblio}
\end{document}